\batchmode
\makeatletter
\def\input@path{{/home/fyzhu/DATA/Dropbox/self_Folder/myWorksOnDropboxs/201702_RLDM_WarmStart_4_onlineRL_based_mHealth/LyxFolders//}}
\makeatother
\documentclass[12pt,preprint, onecolumn]{IEEEtran}
\usepackage[latin9]{inputenc}
\usepackage[a4paper]{geometry}
\usepackage{color}
\definecolor{page_backgroundcolor}{rgb}{1, 1, 1}
\pagecolor{page_backgroundcolor}
\usepackage{array}
\usepackage{bm}
\usepackage{multirow}
\usepackage{amsmath}
\usepackage{setspace}
\usepackage[unicode=true,
 bookmarks=true,bookmarksnumbered=false,bookmarksopen=false,
 breaklinks=false,pdfborder={0 0 1},backref=false,colorlinks=true]
 {hyperref}
\hypersetup{
 citecolor=blue, urlcolor=blue, linkcolor=blue}

\makeatletter

\providecommand{\tabularnewline}{\\}

\usepackage{breakurl}
\usepackage{times}
\usepackage{epsfig}
\usepackage{multirow}
\usepackage{mdwlist}
\usepackage{algorithm}
\usepackage{algorithmic}% 
\usepackage{amsmath}
\usepackage{graphicx} 
\usepackage{url}
\usepackage{graphicx}
\usepackage{breakurl}
\usepackage{epsfig}
\usepackage{multirow}
\usepackage{mdwlist}
\usepackage{algorithm}
\usepackage{algorithmic}% 
\usepackage{enumitem}
\usepackage{url}
\usepackage{array}
\usepackage{float}
\usepackage{amsthm}
\usepackage{amsmath}
\usepackage{mathtools}
\usepackage{amsfonts}
\usepackage{algorithm}
\usepackage{algorithmic}
\usepackage{subfigure}
\usepackage{caption}
\usepackage{epstopdf}
\usepackage{adjustbox}
 \usepackage{amssymb}

\makeatother

\begin{document}
\global\long\def\mtbfA{\mathbf{A}}
 \global\long\def\mtbfa{\mathbf{a}}
 \global\long\def\mebfA{\bar{\mtbfA}}
 \global\long\def\mebfa{\bar{\mtbfa}}

\global\long\def\mhbfA{\widehat{\mathbf{A}}}
 \global\long\def\mhbfa{\widehat{\mathbf{a}}}
 \global\long\def\mtcalA{\mathcal{A}}
 \global\long\def\mtbbA{\mathbb{A}}

\global\long\def\mtbfB{\mathbf{B}}
 \global\long\def\mtbfb{\mathbf{b}}
 \global\long\def\mebfB{\bar{\mtbfB}}
 \global\long\def\mebfb{\bar{\mtbfb}}

\global\long\def\mhbfB{\widehat{\mathbf{B}}}
 \global\long\def\mhbfb{\widehat{\mathbf{b}}}
 \global\long\def\mtcalB{\mathcal{B}}
 \global\long\def\mtbbB{\mathbb{B}}

\global\long\def\mtbfC{\mathbf{C}}
 \global\long\def\mtbfc{\mathbf{c}}
 \global\long\def\mebfC{\bar{\mtbfC}}
 \global\long\def\mebfc{\bar{\mtbfc}}

\global\long\def\mhbfC{\widehat{\mathbf{C}}}
 \global\long\def\mhbfc{\widehat{\mathbf{c}}}
 \global\long\def\mtcalC{\mathcal{C}}
 \global\long\def\mtbbC{\mathbb{C}}

\global\long\def\mtbfD{\mathbf{D}}
 \global\long\def\mtbfd{\mathbf{d}}
 \global\long\def\mebfD{\bar{\mtbfD}}
 \global\long\def\mebfd{\bar{\mtbfd}}

\global\long\def\mhbfD{\widehat{\mathbf{D}}}
 \global\long\def\mhbfd{\widehat{\mathbf{d}}}
 \global\long\def\mtcalD{\mathcal{D}}
 \global\long\def\mtbbD{\mathbb{D}}

\global\long\def\mtbfE{\mathbf{E}}
 \global\long\def\mtbfe{\mathbf{e}}
 \global\long\def\mebfE{\bar{\mtbfE}}
 \global\long\def\mebfe{\bar{\mtbfe}}

\global\long\def\mhbfE{\widehat{\mathbf{E}}}
 \global\long\def\mhbfe{\widehat{\mathbf{e}}}
 \global\long\def\mtcalE{\mathcal{E}}
 \global\long\def\mtbbE{\mathbb{E}}

\global\long\def\mtbfF{\mathbf{F}}
 \global\long\def\mtbff{\mathbf{f}}
 \global\long\def\mebfF{\bar{\mathbf{F}}}
 \global\long\def\mebff{\bar{\mathbf{f}}}

\global\long\def\mhbfF{\widehat{\mathbf{F}}}
 \global\long\def\mhbff{\widehat{\mathbf{f}}}
 \global\long\def\mtcalF{\mathcal{F}}
 \global\long\def\mtbbF{\mathbb{F}}

\global\long\def\mtbfG{\mathbf{G}}
 \global\long\def\mtbfg{\mathbf{g}}
 \global\long\def\mebfG{\bar{\mathbf{G}}}
 \global\long\def\mebfg{\bar{\mathbf{g}}}

\global\long\def\mhbfG{\widehat{\mathbf{G}}}
 \global\long\def\mhbfg{\widehat{\mathbf{g}}}
 \global\long\def\mtcalG{\mathcal{G}}
 \global\long\def\mtbbG{\mathbb{G}}

\global\long\def\mtbfH{\mathbf{H}}
 \global\long\def\mtbfh{\mathbf{h}}
 \global\long\def\mebfH{\bar{\mathbf{H}}}
 \global\long\def\mebfh{\bar{\mathbf{h}}}

\global\long\def\mhbfH{\widehat{\mathbf{H}}}
 \global\long\def\mhbfh{\widehat{\mathbf{h}}}
 \global\long\def\mtcalH{\mathcal{H}}
 \global\long\def\mtbbH{\mathbb{H}}

\global\long\def\mtbfI{\mathbf{I}}
 \global\long\def\mtbfi{\mathbf{i}}
 \global\long\def\mebfI{\bar{\mathbf{I}}}
 \global\long\def\mebfi{\bar{\mathbf{i}}}

\global\long\def\mhbfI{\widehat{\mathbf{I}}}
 \global\long\def\mhbfi{\widehat{\mathbf{i}}}
 \global\long\def\mtcalI{\mathcal{I}}
 \global\long\def\mtbbI{\mathbb{I}}

\global\long\def\mtbfJ{\mathbf{J}}
 \global\long\def\mtbfj{\mathbf{j}}
 \global\long\def\mebfJ{\bar{\mathbf{J}}}
 \global\long\def\mebfj{\bar{\mathbf{j}}}

\global\long\def\mhbfJ{\widehat{\mathbf{J}}}
 \global\long\def\mhbfj{\widehat{\mathbf{j}}}
 \global\long\def\mtcalJ{\mathcal{J}}
 \global\long\def\mtbbJ{\mathbb{J}}

\global\long\def\mtbfK{\mathbf{K}}
 \global\long\def\mtbfk{\mathbf{k}}
 \global\long\def\mebfK{\bar{\mathbf{K}}}
 \global\long\def\mebfk{\bar{\mathbf{k}}}

\global\long\def\mhbfK{\widehat{\mathbf{K}}}
 \global\long\def\mhbfk{\widehat{\mathbf{k}}}
 \global\long\def\mtcalK{\mathcal{K}}
 \global\long\def\mtbbK{\mathbb{K}}

\global\long\def\mtbfL{\mathbf{L}}
 \global\long\def\mtbfl{\mathbf{l}}
 \global\long\def\mebfL{\bar{\mathbf{L}}}
 \global\long\def\mebfl{\bar{\mathbf{l}}}

\global\long\def\mhbfL{\widehat{\mathbf{K}}}
 \global\long\def\mhbfl{\widehat{\mathbf{k}}}
 \global\long\def\mtcalL{\mathcal{L}}
 \global\long\def\mtbbL{\mathbb{L}}

\global\long\def\mtbfM{\mathbf{M}}
 \global\long\def\mtbfm{\mathbf{m}}
 \global\long\def\mebfM{\bar{\mathbf{M}}}
 \global\long\def\mebfm{\bar{\mathbf{m}}}

\global\long\def\mhbfM{\widehat{\mathbf{M}}}
 \global\long\def\mhbfm{\widehat{\mathbf{m}}}
 \global\long\def\mtcalM{\mathcal{M}}
 \global\long\def\mtbbM{\mathbb{M}}

\global\long\def\mtbfN{\mathbf{N}}
 \global\long\def\mtbfn{\mathbf{n}}
 \global\long\def\mebfN{\bar{\mathbf{N}}}
 \global\long\def\mebfn{\bar{\mathbf{n}}}

\global\long\def\mhbfN{\widehat{\mathbf{N}}}
 \global\long\def\mhbfn{\widehat{\mathbf{n}}}
 \global\long\def\mtcalN{\mathcal{N}}
 \global\long\def\mtbbN{\mathbb{N}}

\global\long\def\mtbfO{\mathbf{O}}
 \global\long\def\mtbfo{\mathbf{o}}
 \global\long\def\mebfO{\bar{\mathbf{O}}}
 \global\long\def\mebfo{\bar{\mathbf{o}}}

\global\long\def\mhbfO{\widehat{\mathbf{O}}}
 \global\long\def\mhbfo{\widehat{\mathbf{o}}}
 \global\long\def\mtcalO{\mathcal{O}}
 \global\long\def\mtbbO{\mathbb{O}}

\global\long\def\mtbfP{\mathbf{P}}
 \global\long\def\mtbfp{\mathbf{p}}
 \global\long\def\mebfP{\bar{\mathbf{P}}}
 \global\long\def\mebfp{\bar{\mathbf{p}}}

\global\long\def\mhbfP{\widehat{\mathbf{P}}}
 \global\long\def\mhbfp{\widehat{\mathbf{p}}}
 \global\long\def\mtcalP{\mathcal{P}}
 \global\long\def\mtbbP{\mathbb{P}}

\global\long\def\mtbfQ{\mathbf{Q}}
 \global\long\def\mtbfq{\mathbf{q}}
 \global\long\def\mebfQ{\bar{\mathbf{Q}}}
 \global\long\def\mebfq{\bar{\mathbf{q}}}

\global\long\def\mhbfQ{\widehat{\mathbf{Q}}}
 \global\long\def\mhbfq{\widehat{\mathbf{q}}}
\global\long\def\mtcalQ{\mathcal{Q}}
 \global\long\def\mtbbQ{\mathbb{Q}}

\global\long\def\mtbfR{\mathbf{R}}
 \global\long\def\mtbfr{\mathbf{r}}
 \global\long\def\mebfR{\bar{\mathbf{R}}}
 \global\long\def\mebfr{\bar{\mathbf{r}}}

\global\long\def\mhbfR{\widehat{\mathbf{R}}}
 \global\long\def\mhbfr{\widehat{\mathbf{r}}}
\global\long\def\mtcalR{\mathcal{R}}
 \global\long\def\mtbbR{\mathbb{R}}

\global\long\def\mtbfS{\mathbf{S}}
 \global\long\def\mtbfs{\mathbf{s}}
 \global\long\def\mebfS{\bar{\mathbf{S}}}
 \global\long\def\mebfs{\bar{\mathbf{s}}}

\global\long\def\mhbfS{\widehat{\mathbf{S}}}
 \global\long\def\mhbfs{\widehat{\mathbf{s}}}
\global\long\def\mtcalS{\mathcal{S}}
 \global\long\def\mtbbS{\mathbb{S}}

\global\long\def\mtbfT{\mathbf{T}}
 \global\long\def\mtbft{\mathbf{t}}
 \global\long\def\mebfT{\bar{\mathbf{T}}}
 \global\long\def\mebft{\bar{\mathbf{t}}}

\global\long\def\mhbfT{\widehat{\mathbf{T}}}
 \global\long\def\mhbft{\widehat{\mathbf{t}}}
 \global\long\def\mtcalT{\mathcal{T}}
 \global\long\def\mtbbT{\mathbb{T}}

\global\long\def\mtbfU{\mathbf{U}}
 \global\long\def\mtbfu{\mathbf{u}}
 \global\long\def\mebfU{\bar{\mathbf{U}}}
 \global\long\def\mebfu{\bar{\mathbf{u}}}

\global\long\def\mhbfU{\widehat{\mathbf{U}}}
 \global\long\def\mhbfu{\widehat{\mathbf{u}}}
 \global\long\def\mtcalU{\mathcal{U}}
 \global\long\def\mtbbU{\mathbb{U}}

\global\long\def\mtbfV{\mathbf{V}}
 \global\long\def\mtbfv{\mathbf{v}}
 \global\long\def\mebfV{\bar{\mathbf{V}}}
 \global\long\def\mebfv{\bar{\mathbf{v}}}

\global\long\def\mhbfV{\widehat{\mathbf{V}}}
 \global\long\def\mhbfv{\widehat{\mathbf{v}}}
\global\long\def\mtcalV{\mathcal{V}}
 \global\long\def\mtbbV{\mathbb{V}}

\global\long\def\mtbfW{\mathbf{W}}
 \global\long\def\mtbfw{\mathbf{w}}
 \global\long\def\mebfW{\bar{\mathbf{W}}}
 \global\long\def\mebfw{\bar{\mathbf{w}}}

\global\long\def\mhbfW{\widehat{\mathbf{W}}}
 \global\long\def\mhbfw{\widehat{\mathbf{w}}}
 \global\long\def\mtcalW{\mathcal{W}}
 \global\long\def\mtbbW{\mathbb{W}}

\global\long\def\mtbfX{\mathbf{X}}
 \global\long\def\mtbfx{\mathbf{x}}
 \global\long\def\mebfX{\bar{\mtbfX}}
 \global\long\def\mebfx{\bar{\mtbfx}}

\global\long\def\mhbfX{\widehat{\mathbf{X}}}
 \global\long\def\mhbfx{\widehat{\mathbf{x}}}
 \global\long\def\mtcalX{\mathcal{X}}
 \global\long\def\mtbbX{\mathbb{X}}

\global\long\def\mtbfY{\mathbf{Y}}
 \global\long\def\mtbfy{\mathbf{y}}
\global\long\def\mebfY{\bar{\mathbf{Y}}}
 \global\long\def\mebfy{\bar{\mathbf{y}}}

\global\long\def\mhbfY{\widehat{\mathbf{Y}}}
 \global\long\def\mhbfy{\widehat{\mathbf{y}}}
 \global\long\def\mtcalY{\mathcal{Y}}
 \global\long\def\mtbbY{\mathbb{Y}}

\global\long\def\mtbfZ{\mathbf{Z}}
 \global\long\def\mtbfz{\mathbf{z}}
 \global\long\def\mebfZ{\bar{\mathbf{Z}}}
 \global\long\def\mebfz{\bar{\mathbf{z}}}

\global\long\def\mhbfZ{\widehat{\mathbf{Z}}}
 \global\long\def\mhbfz{\widehat{\mathbf{z}}}
\global\long\def\mtcalZ{\mathcal{Z}}
 \global\long\def\mtbbZ{\mathbb{Z}}

\global\long\def\mtth{\text{th}}

\global\long\def\mtbfzero{\mathbf{0}}
 \global\long\def\mtbfone{\mathbf{1}}

\global\long\def\mttrace{\text{Tr}}

\global\long\def\mttotalVariation{\text{TV}}

\global\long\def\mtexpect{\mathbb{E}}

\global\long\def\mtdet{\text{det}}

\global\long\def\mtvec{\mathbf{\text{vec}}}

\global\long\def\mtvar{\mathbf{\text{var}}}

\global\long\def\mtcov{\mathbf{\text{cov}}}

\global\long\def\mtsubTo{\mathbf{\text{s.t.}}}

\global\long\def\mtfor{\text{for}}

\global\long\def\mtrank{\text{rank}}

\global\long\def\mtrankn{\text{rankn}}

\global\long\def\mtdiag{\mathbf{\text{diag}}}

\global\long\def\mtsign{\mathbf{\text{sign}}}

\global\long\def\mtloss{\mathbf{\text{loss}}}

\global\long\def\mtwhen{\text{when}}

\global\long\def\mtwhere{\text{where}}

\global\long\def\mtif{\text{if}}

\title{Effective Warm Start for the Online Actor-Critic Reinforcement Learning
based mHealth Intervention\\
}

\author{\IEEEauthorblockN{Feiyun Zhu\IEEEauthorrefmark{1}\IEEEauthorrefmark{2} and Peng Liao\IEEEauthorrefmark{2}}\\
\IEEEauthorblockA{\IEEEauthorrefmark{1}Department of Statistic, Univeristy of Michigan,\\
1085 S University Ave, Ann Arbor, MI 4810} \\
\IEEEauthorblockA{\IEEEauthorrefmark{2}Department of Computer Science \& Engineering,
University of Texas at Arlington\\
500 UTA Boulevard, Arlington, TX 76019-0015}}
\maketitle
\begin{abstract}
Online reinforcement learning (RL) is increasingly popular for the
personalized mobile health (mHealth) intervention. It is able to personalize
the type and dose of interventions according to user's ongoing statuses
and changing needs. However, at the beginning of online learning,
there are usually too few samples to support the RL updating, which
leads to poor performances. A delay in good performance of the online
learning algorithms can be especially detrimental in the mHealth,
where users tend to quickly disengage with the mHealth app. To address
this problem, we propose a new online RL methodology that focuses
on an effective warm start. The main idea is to make full use of the
data accumulated and the decision rule achieved in a former study.
As a result, we can greatly enrich the data size at the beginning
of online learning in our method. Such case accelerates the online
learning process for new users to achieve good performances not only
at the beginning of online learning but also through the whole online
learning process. Besides, we use the decision rules achieved in a
previous study to initialize the parameter in our online RL model
for new users. It provides a good initialization for the proposed
online RL algorithm. Experiment results show that promising improvements
have been achieved by our method compared with the state-of-the-art
method.\end{abstract}


\section{Introduction}

\IEEEPARstart{W}{ith} billions of smart device (i.e., smart-phones
and wearable devices) users worldwide, mobile health (mHealth) interventions
(MHI) are increasingly popular among the behavioral health, clinical,
computer science and statistic communities\ \cite{huitian_2014_NIPS_ActCriticBandit4JITAI,SusanMurphy_2016_CORR_BatchOffPolicyAvgRwd,Walter_2015_Significance_RandomTrialForFitbitGeneration,PengLiao_2015_Proposal_offPolicyRL}.
The MHI aims to make full use of smart technologies to collect, transport
and analyze the raw data (weather, location, social activity, stress,
urges to smoke, etc.) to deliver effective treatments that target
behavior regularization\ \cite{SusanMurphy_2016_CORR_BatchOffPolicyAvgRwd}.
For example, the goal of MHI is to optimally prevent unhealthy behaviors,
such as alcohol abuse and eating disorders, and to promote healthy
behaviors. Particularly, JITAIs (i.e., Just in time adaptive intervention)
is especially interesting and practical due to the appealing properties\ \cite{huitian_2014_NIPS_ActCriticBandit4JITAI}:
(1) JITAIs could make adaptive and efficacious interventions according
to user's ongoing statuses and changing needs; (2) JITAIs allow for
the real-time delivery of interventions, which is very portable, affordable
and flexible\ \cite{fyZhu_2017_arXiv_CohesionDrivenActorCriticRL}.
Therefore, JITAIs are widely used in a wide range of mHealth applications,
such as physical activity, eating disorders, alcohol use, mental illness,
obesity/weight management and other chronic disorders etc., that aims
to guide people to lead healthy lives\ \cite{PengLiao_2015_Proposal_offPolicyRL,SusanMurphy_2016_CORR_BatchOffPolicyAvgRwd,huitian_2016_PhdThesis_actCriticAlgorithm,Walter_2015_Significance_RandomTrialForFitbitGeneration,Gustafson_2014_JAMA_drinking}.

Normally, JITAIs is formed as an online sequential decision making
(SDM) problem that is aimed to construct the optimal decision rules
to decide when, where and how to deliver effective treatments\ \cite{PengLiao_2015_Proposal_offPolicyRL,SusanMurphy_2016_CORR_BatchOffPolicyAvgRwd,fyZhu_2017_arXiv_CohesionDrivenActorCriticRL}.
This is a brand-new topic that lacks of methodological guidance. In
2014, Lei\ \cite{huitian_2014_NIPS_ActCriticBandit4JITAI} made a
first attempt to formulate the mHealth intervention as an online actor-critic
contextual bandit problem. Lei's method is well suited for the small
data set problem in the early stage of the mHealth study. However,
this method ignores the important delayed effects of the SDM---the
current action may affect not only the immediate reward but also the
next states and, through that, all subsequent rewards\ \cite{Sutton_2012_Book_ReinforcementLearning}.
To consider the delayed effects, it is reasonable to employ the reinforcement
learning (RL) in the discount reward setting. RL is much more complex
than the contextual bandit. It requires much more data to acquire
good and stable decision rules\ \cite{fyZhu_2017_arXiv_CohesionDrivenActorCriticRL}.
However at the beginning of the online learning, there are too few
data to start effective online learning. A simple and widely used
method is to collect a fixed length of trajectory ($T_{0}=10,$ say)
via the micro-randomized trials\ \cite{PengLiao_2015_Proposal_offPolicyRL},
accumulating a few of samples, then starting the online updating.
Such procedure is called the random warm start, i.e. RWS. However,
there are two main drawbacks of the RWS: (1) it highly puts off the
online RL learning before achieving good decision rules; (2) it is
likely to frustrate the users because the random decision rules achieved
at the beginning of online learning are not personalized to the users'
needs. Accordingly, it is easy for users to abandon the mHealth app. 

To alleviate the above problems, we propose a new online RL methodology
by emphasizing effective warm starts. It aims to promote the performance
of the online learning  at the early stage and, through that, the
final decision rule. The motivation is to make full use of the data
and the decision rules achieved in the previous study, which is similar
to the current study (cf. Sec.\ \ref{sec:OurMethods}). Specifically,
we use the decision rules achieved previously to initialize the parameter
of the online RL learning for new users. The data accumulated in the
former study is also fully used. As a result, the data size is greatly
enriched at the beginning of our online learning algorithm. When the
online learning goes on, the data gained from new users will have
more and more weights to increasingly dominate the objective function.
Our decision rule is still personalized according to the new user.
Extensive experiment results show the power of the proposed method.

\section{Markov Decision Process (MDP) and Actor-Critic Reinforcement Learning}

\textbf{MDP}: The dynamic system (i.e. the environment) that RL interacts
with is generally modeled as a Markov Decision Process (MDP)\ \cite{Sutton_2012_Book_ReinforcementLearning}.
An MDP is a tuple $\left\{ \mtcalS,\mtcalA,P,R,\gamma\right\} $\ \cite{Geist_2013_TNNLS_RL_valueFunctionApproximation,Grondman_2012_IEEEts_surveyOfActorCriticRL,Michail_2003_JMLR_LSPI_LSTDQ},
where $\mtcalS$ is (finite) state space and $\mtcalA$ is (finite)
action space. The state transition probability $P:\mtcalS\times\mtcalA\times\mtcalS\mapsto\left[0,1\right]$,
from state $s$ to the next state $s'$ when taking action $a$, is
given by $P(s,a,s')$. Let $S_{t},A_{t}$ and $R_{t+1}$ be the random
variables at time $t$ representing the state, action and immediate
reward respectively. The expected immediate reward $R\left(s,a\right)=\mtexpect\left(R_{t+1}\mid S_{t}=s,A_{t}=a\right)$
is assumed to be bounded over the state and action spaces\ \cite{SusanMurphy_2016_CORR_BatchOffPolicyAvgRwd}.
$\gamma\in[0,1)$ is a discount factor to reduce the influence of
future rewards.

The stochastic policy $\pi\left(\cdot\mid s\right)$ decides the action
to take in a given state $s$. The goal of RL is to interact with
the environment  to learn the optimal policy $\pi^{*}$ that maximizes
the total accumulated reward. Usually, RL uses the value function
$Q^{\pi}\left(s,a\right)\in\mtbbR^{\left|\mtcalS\right|\times\left|\mtcalA\right|}$
to quantify the quality of a policy $\pi$, which is the expected
discounted cumulative reward, starting from state $s$, first choosing
action $a$ and then following the policy $\pi$: $Q^{\pi}\left(s,a\right)=\mtexpect\left\{ \sum_{t=0}^{\infty}\gamma^{t}R\left(s_{t},a_{t}\right)\mid s_{0}=s,a_{0}=a,\pi\right\} .$
The value $Q^{\pi}\left(s,a\right)$ satisfies the following linear
Bellman equation 
\begin{equation}
Q^{\pi}\left(s,a\right)=\mtexpect_{s',a'\mid s,a,\pi}\left\{ R\left(s,a\right)+\gamma Q^{\pi}\left(s',a'\right)\right\} \label{eq:Q_BellmanEqn}
\end{equation}
The parameterized functions are generally employed to approximate
the value and policy functions since\ \cite{Geist_2013_TNNLS_RL_valueFunctionApproximation}
the system usually have too many states and actions to achieve an
accurate estimation of value and policy. Instead they have to be iteratively
estimated. Due to the great properties of quick convergences\ \cite{Grondman_2012_IEEEts_surveyOfActorCriticRL},
the actor-critic RL algorithms are widely accepted to esimate the
parameterized value $Q_{\mtbfw}\left(s,a\right)=\mtbfw^{T}\mtbfx\left(s,a\right)\approx Q^{\pi}$
and stochastic policy $\pi_{\theta}\left(\cdot\mid s\right)\approx\pi^{*}\left(\cdot\mid s\right)$,
where $\mtbfx\left(s,a\right)$ is a feature function for the $Q$-value
that merges the information in state $s$ and action $a$. To learn
the unknown parameters $\left\{ \mtbfw,\theta\right\} $, we need
a 2-step alternating updating rule until convergence: (1) the critic
updating (i.e., policy evaluation) for $\mtbfw$ to estimate the Q-value
function for the current policy, (2) the actor updating (i.e., policy
improvement) for $\theta$ to search a better policy based on the
newly estimated Q-value\ \cite{Grondman_2012_IEEEts_surveyOfActorCriticRL,PengLiao_2015_Proposal_offPolicyRL}.

Supposing the online learning for a new user is at decision point
$t$, resulting in $t$ tuples drawn from the MDP system, i.e., $\mtcalD=\left\{ \left(s_{i},a_{i},r_{i},s'_{i}\right)\mid i=1,\cdots,t\right\} $.
Each tuple consists of four elements: the current state, action, reward
and the next state$.$ By using the data in $\mtcalD,$ the Least-Squares
Temporal Difference for Q-value (LSTD$Q$)\ \cite{Michail_2003_JMLR_LSPI_LSTDQ,Sutton_2012_Book_ReinforcementLearning}
is used for the\textbf{\emph{ }}\emph{critic updating} to estimate
\emph{$\widehat{\mathbf{w}}_{t}$ }at time point $t$:
\begin{equation}
\widehat{\mathbf{w}}_{t}=\left[\zeta_{c}\mathbf{I}+\frac{1}{t}\sum_{i=1}^{t}\mathbf{x}_{i}\left(\mathbf{x}_{i}-\gamma\mathbf{y}_{i+1}\right)^{\intercal}\right]^{-1}\left(\frac{1}{t}\sum_{i=1}^{t}\mathbf{x}_{i}r_{i}\right),\label{eq:eq:current_Critic_obj}
\end{equation}
where $\mathbf{x}_{i}=\mathbf{x}\left(s_{i},a_{i}\right)$ is the
feature at decision point $i$ for the value function; 
\[
\mathbf{y}_{i+1}=\sum_{a\in\mtcalA}\mathbf{x}\left(s_{i+1},a\right)\pi_{\hat{\theta}_{t}}\left(a\mid s_{i+1}\right)
\]
 is the feature at the next time point; $r_{i}$ is the immediate
reward at the $i^{\mtth}$ time point\emph{. }By maximizing the average
reward, i.e., a widely accepted criterion\ \cite{Grondman_2012_IEEEts_surveyOfActorCriticRL},
we have the objective function for the \emph{actor updating} (i.e.,
policy improvement)\ \emph{
\begin{equation}
\widehat{\theta}_{t}=\arg\max_{\theta}\ \frac{1}{t}\sum_{i=1}^{t}\sum_{a\in\mtcalA}Q\left(s_{i},a;\mathbf{\mhbfw}_{t}\right)\pi_{\theta}\left(a|s_{i}\right)-\frac{\zeta_{a}}{2}\left\Vert \theta\right\Vert _{2}^{2}\label{eq:current_actor_obj}
\end{equation}
}where $Q\left(s_{i},a;\widehat{\mathbf{w}}_{t}\right)\!=\!\mathbf{x}\left(s_{i},a\right)^{\intercal}\widehat{\mathbf{w}}_{t}$
is the newly estimated value; $\zeta_{c}$ and $\zeta_{a}$ are the
balancing parameters for the $\ell_{2}$ constraint to avoid singular
failures for the critic and actor update respectively. Note that after
each actor update, the feature at the next time point $\mathbf{y}_{i+1}$
has to be re-calculated based on the newly estimated policy parameter
$\widehat{\theta}_{t}.$ When the discount factor $\gamma=0$, the
RL algorithm in\ \eqref{eq:our_Critic_obj},\ \eqref{eq:our_Actor_obj}
is equivalent to the state-of-the-art contextual bandit method in
the mHealth\ \cite{huitian_2014_NIPS_ActCriticBandit4JITAI}.

\section{\label{sec:OurMethods}Our method}

The actor-critic RL algorithm in\ \eqref{eq:our_Critic_obj},\ \eqref{eq:our_Actor_obj}
works well when the sample size (i.e. $t$) is large. However at the
beginning of the online learning, e.g., $t=1,$ there is only one
tuple. It is impossible to do the actor-critic updating with so few
samples. A popular and widely accepted method is to accumulate a small
number of tuples via the micro-randomized trials\ \cite{PengLiao_2015_Proposal_offPolicyRL}
(called RWS). RWS is to draw a fixed length of trajectory ($T_{0}=10$,
say) by applying the random policy with probability 0.5 to provide
an intervention (i.e., $\mu\left(1\mid s\right)=0.5$ for all states
$s$). RWS works to some extent, they are far from the optimal. One
direct drawback with RWS is that it is very expensive in time to wait
the micro-randomized trials to collect data from human, implying that
we may still have a small number of samples to start the actor-critic
updating. This tough problem badly affects the actor-critic updating
not only at the beginning of online learning, but also along the whole
learning process. Such case is due to the actor-critic objective functions
is non-convex; any bad solution at the early online learning would
bias the optimization direction, which easily leads some sub-optimal
solution. Besides, the random policy in micro-randomized trials and
the decision rules achieved at the early online learning is of bad
user experience. Such problem makes it possible for the users to be
inactive with or even to abandon the mHealth intervention.

To deal with the above problems, we propose a new online actor-critic
RL methodology. It emphasizes effective warm starts for the online
learning algorithm. The goal is to promote decision rules achieved
at the early online learning stage and, through that, guide the optimization
in a better direction, leading to a good final policy that is well
suited for the new user. Specifically, we make full use of the data
accumulated and decision rules learned in the previous study. Note
that for the mHealth intervention design, there are usually several
rounds of study; each round is pushed forward and slightly different
from the former one. By using the data and policy gained in the former
study, the RL learning in current study could quickly achieve good
decision rules for new users, reducing the total study time and increasing
the user experience at the beginning of the online learning.

Supposing that the former mHealth study is carried out in an off-policy,
batch learning setting, we have $\bar{N}$ (40, say) individuals.
Each individual is with a trajectory including $\bar{T}=42$ tuples
of states, actions and rewards. Thus in total there are $NT=\bar{N}\times\bar{T}$
tuples, i.e., $\bar{\mtcalD}=\left\{ \left(\bar{s}_{i},\bar{a}_{i},\bar{r}_{i},\bar{s}'_{i}\right)\mid i=1,\cdots,NT\right\} $.
Besides the data in $\bar{\mtcalD}$, we employ the decision rule
achieved in the former study to initialize the parameters in the current
online learning. Note that we add a bar above the notations to distinguish
the data obtained in the previous study from that of the current study.

At the $t^{\mtth}$ decision point, we have both the data $\bar{\mtcalD}$
collected in the former study and the $t$ new tuples drawn from the
new user in $\mtcalD$ to update the online actor-critic learning.
It has two parts: (1) the \emph{critic updating}\textbf{\emph{ }}for\emph{
$\widehat{\mathbf{w}}_{t}$ }via
\begin{align}
\widehat{\mathbf{w}}_{t}= & \left\{ \zeta_{c}\mathbf{I}+\frac{1}{t+1}\left[{\color{blue}\frac{1}{NT}\sum_{j=1}^{NT}\mathbf{\mebfx}_{j}\left(\mathbf{\mebfx}_{j}-\gamma\mathbf{\mebfy}_{i+1}\right)^{\intercal}}+\sum_{i=1}^{t}\mathbf{x}_{i}\left(\mathbf{x}_{i}-\gamma\mathbf{y}_{i+1}\right)^{\intercal}\right]\right\} ^{-1}\label{eq:our_Critic_obj}\\
 & \left[\frac{1}{t+1}\left({\color{blue}\frac{1}{NT}\sum_{j=1}^{NT}\mebfx_{j}\bar{r}_{j}}+\sum_{i=1}^{t}\mathbf{x}_{i}r_{i}\right)\right]\nonumber 
\end{align}
and (2) the \emph{actor updating}\textbf{ }via 
\begin{align}
\widehat{\theta}_{t}=\arg\max_{\theta}\  & \frac{1}{t+1}\left\{ {\color{blue}\frac{1}{NT}\sum_{j=1}^{NT}\sum_{a\in\mtcalA}Q\left(\bar{s}_{j},a;\widehat{\mathbf{w}}_{t}\right)\pi_{\theta}\left(a|\bar{s}_{j}\right)}+\sum_{i=1}^{t}\sum_{a\in\mtcalA}Q\left(s_{i},a;\widehat{\mathbf{w}}_{t}\right)\cdot\pi\left(a|s_{i}\right)\right\} -\frac{\zeta_{a}}{2}\left\Vert \theta\right\Vert _{2}^{2},\label{eq:our_Actor_obj}
\end{align}
where $\left\{ \mebfx_{j}\right\} _{j=1}^{NT}$ is data in the previous
study; $\left(\mtbfx_{i}\right)_{i=1}^{t}$ is the data that is collected
from the new user; $\mebfx_{i}=\mathbf{\mebfx}\left(s_{i},a_{i}\right)$
is the feature vector at decision point $i$ for the value function;
$\mebfy_{i+1}=\sum_{a\in\mtcalA}\mathbf{x}\left(\bar{s}_{i+1},a\right)\pi_{\hat{\theta}_{t}}\left(a\mid\bar{s}_{i+1}\right)$
is the feature at the next time point; $\bar{r}_{i}$ is the immediate
reward at the $i^{\mtth}$ point; $Q\left(\bar{s}_{i},a;\widehat{\mathbf{w}}_{t}\right)=\mathbf{x}\left(\bar{s}_{i},a\right)^{\intercal}\widehat{\mathbf{w}}_{t}$
is the newly updated value.

In\ \eqref{eq:our_Critic_obj} and\ \eqref{eq:our_Actor_obj}, the
terms in the blue ink indicate the the previous data, which is with
a normalized weight $\frac{1}{NT}.$ In this setting, all the data
obtained in the former study is treated as one sample for the current
online learning. When current online learning goes on (i.e., $t$
increases), the data collected from the new user gradually dominates
the objective functions. Thus, we are still able to achieve personalized
JITAIs that is successfully adapted to each new user.

\section{Experiments}

To verify the performance, we compare our method (i.e., NWS-RL) with
the conventional RL method with the random warm start (RWS-RL) on
the HeartSteps application\ \cite{Walter_2015_Significance_RandomTrialForFitbitGeneration}.
The HeartSteps is a 42-day mHealth intervention that encourages users
to increase the steps they take each day by providing positive interventions,
such as suggesting taking a walk after sedentary behavior\textbf{.}
The actions are binary including $\left\{ 0,1\right\} $, where $a=1$
means providing active treatments, e.g., sending an intervention to
the user's smart device, while $a=0$ means no treatment\ \cite{SusanMurphy_2016_CORR_BatchOffPolicyAvgRwd}.

\subsection{Simulated Experiments}

In the experiments, we draw $T$ tuples from each user, i.e., 
\[
\mathcal{D}_{T}=\left\{ \left(S_{0},A_{0},R_{0}\right),\left(S_{1},A_{1},R_{1}\right),\cdots,\left(S_{T},A_{T},R_{T}\right)\right\} 
\]
, where the observation $S_{t}$ is a column vector with $p$ elements
. The initial states and actions are generated by $S_{0}\sim\mathrm{Normal}_{p}\left\lbrace 0,\Sigma\right\rbrace $
and $A_{0}=0$, where $\Sigma=\left[\begin{array}{cc}
\Sigma_{1} & 0\\
0 & I_{p-3}
\end{array}\right]$ and ${\displaystyle \Sigma_{1}=\begin{bmatrix}1 & 0.3 & -0.3\\
0.3 & 1 & -0.3\\
-0.3 & -0.3 & 1
\end{bmatrix}}.$ For $t\geq1$, we have the state generation model and immediate reward
model as follows
\begin{align}
S_{t,1} & =\beta_{1}S_{t-1,1}+\xi_{t,1},\nonumber \\
S_{t,2} & =\beta_{2}S_{t-1,2}+\beta_{3}A_{t-1}+\xi_{t,2},\label{eq:Dat=0000231_stateTrans_cmp3}\\
S_{t,3} & =\beta_{4}S_{t-1,3}+\beta_{5}S_{t-1,3}A_{t-1}+\beta_{6}A_{t-1}+\xi_{t,3},\nonumber \\
S_{t,j} & =\beta_{7}S_{t-1,j}+\xi_{t,j},\quad\mtfor\ j=4,\ldots,p\nonumber \\
R_{t} & =\beta_{14}\times\left[\beta_{8}+A_{t}\times(\beta_{9}+\beta_{10}S_{t,1}+\beta_{11}S_{t,2})+\beta_{12}S_{t,1}-{\color{blue}\beta_{13}S_{t,3}}+\varrho_{t}\right],\label{eq:Dat=0000231_ImmediateRwd_cmp3}
\end{align}
where $-\beta_{13}O_{t,3}$ is the treatment fatigue\ \cite{PengLiao_2015_Proposal_offPolicyRL,SusanMurphy_2016_CORR_BatchOffPolicyAvgRwd};
$\left\{ \xi_{t,i}\right\} _{i=1}^{p}\sim\textnormal{Normal}\left(0,\sigma_{s}^{2}\right)$
at the $t^{\mtth}$ point is the noise in the state transition\ \eqref{eq:Dat=0000231_stateTrans_cmp3}
and $\varrho_{t}\sim\textnormal{Normal}\left(0,\sigma_{r}^{2}\right)$
is the noise in the immediate reward model\ \eqref{eq:Dat=0000231_ImmediateRwd_cmp3}.
To generate $N$ different users, we need $N$ different MDPs specified
by the value of $\bm{\beta}$s in\ \eqref{eq:Dat=0000231_stateTrans_cmp3}
and\ \eqref{eq:Dat=0000231_ImmediateRwd_cmp3}. The $\bm{\beta}s$
are generated in the following two steps: (a) set a basic $\bm{\beta}_{\text{basic}}=\left[0.40,0.25,0.35,0.65,0.10,0.50,0.22,2.00,0.15,0.20,0.32,0.10,0.45,1.50,800\right]$;
(b) to obtain $N$ different $\bm{\beta}$s (i.e., users or MDPs),
we set the $\left\{ \bm{\beta}_{i}\right\} _{i=1}^{N}$ as $\bm{\beta}_{i}=\bm{\beta}_{\text{basic}}+\bm{\delta}_{i},\ \text{for}\ i\in\left\{ 1,2,\cdots,N\right\} ,$
where $\bm{\delta}_{i}\sim\text{Normal}\left(0,\sigma_{b}\mathbf{I}_{14}\right)$,
$\sigma_{b}$ controls how different the users are and $\mathbf{I}_{14}$
is an identity matrix with $14\times14$ elements. To generate the
MDP for a future user, we will also use this kind of method to generate
new $\bm{\beta}$s.

\subsection{Experiment Settings}

The expectation of long run average reward (ElrAR) $\mathbb{E}\left[\eta^{\pi_{\widehat{\theta}}}\right]$
is used to evaluate the quality of an estimated policy $\pi_{\widehat{\theta}}$
on a set of $N$=50 individuals. Intuitively, the ElrAR measures how
much average reward in the long run we could totally get by using
the learned policy $\pi_{\widehat{\theta}}$ for a number of users.
In the HeartSteps application, the ElrAR measures the average steps
that users take each day in a long period of time; a larger ElrAR
corresponds to a better performance. The average reward $\eta^{\pi_{\widehat{\theta}}}$
is calculated by averaging the rewards over the last $4,000$ elements
in a trajectory of $5,000$ tuples under the policy $\pi_{\widehat{\theta}}$.
Then ElrAR $\mathbb{E}\left[\eta^{\pi_{\widehat{\theta}}}\right]$
is achieved by averaging the $50$ $\eta^{\pi_{\widehat{\theta}}}$'s. 

In the experiment, we assume the parameterized policy in the form
\[
\pi_{\theta}\left(a\mid s\right)=\exp\left[a\theta^{\intercal}\phi\left(s\right)\right]/\left\{ 1+\exp\left[\theta^{\intercal}\phi\left(s\right)\right]\right\} 
\]
, where $\theta\in\mtbbR^{q}$ is the unknown variance and $\phi\left(s\right)=\left[1,s^{\intercal}\right]^{\intercal}\in\mtbbR^{q}$
is the feature function for policies that stacks constant 1 with the
state vector $s$. The number of individuals in the former study is
$\bar{N}=40$. Each is with a trajectory of $\bar{T}$=42 time points.
For the current study (cf., Table\ \ref{tab:AverageRwd_RWS_NWS}),
there are $N=50$ individuals. RWS has to accumulate tuples till $T_{0}=5$
and $10$ respectively to start the online learning. Our method (i.e.,
NWS) has the ability to start the RL online learning algorithm immediately
when the $1^{\text{st}}$ tuple is available. Since the comparison
of early online learning is our focuses, we set the total trajectory
length for the online learning as $T=30$ and $T=50$, respectively.
The noises are set $\sigma_{s}=\sigma_{r}=1$ and $\sigma_{\beta}=0.005$.
Other variances are $p=3$, $q=4$, $\zeta_{a}=\zeta_{c}=10^{-5}$.
The feature processing for the value estimation is $\mtbfx\left(s,a\right)=\left[1,s^{\intercal},a,s^{\intercal}a\right]^{\intercal}\in\mtbbR^{2p+2}$
for all the compared methods. Table\ \ref{tab:AverageRwd_RWS_NWS}
summarizes the experiment results of three methods: RWS-RL$_{T_{0}=5}$,
RWS-RL$_{T_{0}=10}$ and NWS-RL$_{T_{0}=1}$. It includes two sub-tables:
the left one shows the results of early online learning results, i.e.,
$T=30$ and the right displays the results when $T=50$. As we shall
see, the proposed warm start method (NWS-RL) has an obvious advantage
over the conventional RWS-RL, averagely achieving an improvement of
$67.57$ steps for $T=30$ and $69.72$ steps for $T=50$ compared
with the $2^{\text{nd}}$ best policy in \textit{\textcolor{blue}{blue}}.
\begin{table}
\begin{centering}
\caption{The average reward of three online RL methods as discount factor $\gamma$
rises from $0$ to $0.95$: (a) Random Warm Start RL (RWS-RL) when
$T_{0}=5$ and $T_{0}=10$ respectively; (b) the proposed New Warm
Start RL (NWS-RL) is able to start the online learning when the $1^{\text{st}}$
tuple is available, i.e., $T_{0}=1$. The\textbf{\textcolor{red}{{}
Red value}} is the best and the \textcolor{blue}{\emph{blue value}}
is the $2^{\text{nd}}$ best. \label{tab:AverageRwd_RWS_NWS}}
\begin{adjustbox}{width=0.99\linewidth, center}%
\begin{tabular}{|c|c|c|c||c|c|c|}
\hline 
\multirow{2}{*}{$\gamma$ } &
\multicolumn{3}{c||}{Average reward when trajectory length $T=30$ } &
\multicolumn{3}{c|}{Average reward when trajectory length $T=50$ }\tabularnewline
\cline{2-7} 
 & RWS-RL$_{T_{0}=5}$ &
RWS-RL$_{T_{0}=10}$ &
NWS-RL$_{T_{0}=1}$ &
RWS-RL$_{T_{0}=5}$ &
RWS-RL$_{T_{0}=10}$ &
NWS-RL$_{T_{0}=1}$\tabularnewline
\hline 
$0$ &
1152.9$\pm$18.8 &
\textit{\textcolor{blue}{1349.7$\pm$13.9 }} &
\textbf{\textcolor{red}{1367.6$\pm$8.30}} &
1152.6$\pm$19.8 &
\textit{\textcolor{blue}{1335.8$\pm$8.50}} &
\textbf{\textcolor{red}{1365.2$\pm$8.30}}\tabularnewline
$0.2$ &
1152.4$\pm$22.2 &
\textit{\textcolor{blue}{1320.2$\pm$26.8}} &
\textbf{\textcolor{red}{1339.3$\pm$13.6}} &
1153.8$\pm$19.9 &
\textit{\textcolor{blue}{1325.4$\pm$14.2}} &
\textbf{\textcolor{red}{1361.4$\pm$8.70}}\tabularnewline
$0.4$ &
1149.1$\pm$23.8 &
\textit{\textcolor{blue}{1300.6$\pm$32.6}} &
\textbf{\textcolor{red}{1337.7$\pm$23.9 }} &
1149.7$\pm$17.5 &
\textit{\textcolor{blue}{1308.0$\pm$19.7}} &
\textbf{\textcolor{red}{1335.1$\pm$15.8}}\tabularnewline
$0.6$ &
1155.2$\pm$29.0 &
\textit{\textcolor{blue}{1301.1$\pm$32.7}} &
\textbf{\textcolor{red}{1405.4$\pm$49.1}} &
1160.2$\pm$21.3 &
\textit{\textcolor{blue}{1281.2$\pm$27.3}} &
\textbf{\textcolor{red}{1387.6$\pm$48.3}}\tabularnewline
$0.8$ &
1267.8$\pm$43.3 &
\textit{\textcolor{blue}{1326.7$\pm$29.8}} &
\textbf{\textcolor{red}{1481.9$\pm$31.6}} &
1263.0$\pm$37.1 &
\textit{\textcolor{blue}{1333.3$\pm$43.1}} &
\textbf{\textcolor{red}{1501.2$\pm$35.5}}\tabularnewline
$0.95$ &
\multicolumn{1}{c|}{1327.9$\pm$46.0} &
\textit{\textcolor{blue}{1354.9$\pm$27.5}} &
\textbf{\textcolor{red}{1426.7$\pm$30.6}} &
1320.8$\pm$47.2 &
\textit{\textcolor{blue}{1427.7$\pm$33.7}} &
\textbf{\textcolor{red}{1479.3$\pm$39.7}}\tabularnewline
\hline 
\emph{Avg.} &
1200.9 &
\textit{\textcolor{blue}{1325.5}} &
\textbf{\textcolor{red}{1393.1}} &
1200.0 &
\textit{\textcolor{blue}{1335.2}} &
\textbf{\textcolor{red}{1405.0}}\tabularnewline
\hline 
\end{tabular}\end{adjustbox} \vspace{0.1cm}
\par\end{centering}

\emph{\footnotesize{}The value of $\gamma$ specifies different RL
methods: (a) $\gamma=0$ means the contextual bandit\ \cite{huitian_2014_NIPS_ActCriticBandit4JITAI},
(b) $0<\gamma<1$ indicates the discounted reward RL. }{\footnotesize \par}
\end{table}

\section{Conclusion and \emph{Discussion}}

In this paper, we propose a new online actor-critic reinforcement
learning methodology for the mHealth application. The main idea is
to provide an effective warm start method for the online RL learning.
The state-of-the-art RL method for mHealth has the problem of lacking
samples to start the online learning. To solve this problem, we make
full use of the data accumulated and decision rules achieved in the
former study. As a result, the data size is greatly enriched even
at the beginning of online learning. Our method is able to start the
online updating when the first tuple is available. Experiment results
verify that our method achieves clear gains compared with the state-of-the-art
method. In the future, we may explore the robust learning\ \cite{fyzhu_2014_AAAI_ARSS,yingWang_2015_TIP_RobustUnmixing}
and graph learning\ \cite{fyzhu_2014_IJPRS_SSNMF,haichangLi_2016_IJRS_LablePropagationHyperClassification}
on the online actor-critic RL learning algorithm.

\section*{Acknowledgements }

The authors would like to thank the editor and the reviewers for their
valuable suggestions. Besides, this work is supported by R01 AA023187,
P50 DA039838, U54EB020404, R01 HL125440.

{\footnotesize{}\bibliographystyle{ieeetr}
\phantomsection\addcontentsline{toc}{section}{\refname}\bibliography{0_home_fyzhu_link2dropbox_self_Folder_myWorksOnDropboxs_bibFiles_referenceBib}

\begin{thebibliography}{10}

\bibitem{huitian_2014_NIPS_ActCriticBandit4JITAI}
H.~Lei, A.~Tewari, and S.~Murphy, ``An actor-critic contextual bandit algorithm
  for personalized interventions using mobile devices,'' in {\em NIPS 2014
  Workshop: Personalization: Methods and Applications}, pp.~1 -- 9, 2014.

\bibitem{SusanMurphy_2016_CORR_BatchOffPolicyAvgRwd}
S.~A. Murphy, Y.~Deng, E.~B. Laber, H.~R. Maei, R.~S. Sutton, and
  K.~Witkiewitz, ``A batch, off-policy, actor-critic algorithm for optimizing
  the average reward,'' {\em CoRR}, vol.~abs/1607.05047, 2016.

\bibitem{Walter_2015_Significance_RandomTrialForFitbitGeneration}
W.~Dempsey, P.~Liao, P.~Klasnja, I.~Nahum-Shani, and S.~A. Murphy, ``Randomised
  trials for the fitbit generation,'' {\em Significance}, vol.~12, pp.~20 --
  23, Dec 2016.

\bibitem{PengLiao_2015_Proposal_offPolicyRL}
P.~Liao, A.~Tewari, and S.~Murphy, ``Constructing just-in-time adaptive
  interventions,'' {\em Phd Section Proposal}, pp.~1--49, 2015.

\bibitem{fyZhu_2017_arXiv_CohesionDrivenActorCriticRL}
F.~Zhu, P.~Liao, X.~Zhu, Y.~Yao, and J.~Huang, ``Cohesion-based online
  actor-critic reinforcement learning for mhealth intervention,'' {\em
  arXiv:1703.10039}, 2017.

\bibitem{huitian_2016_PhdThesis_actCriticAlgorithm}
H.~Lei, {\em An Online Actor Critic Algorithm and a Statistical Decision
  Procedure for Personalizing Intervention}.
\newblock PhD thesis, University of Michigan, 2016.

\bibitem{Gustafson_2014_JAMA_drinking}
D.~Gustafson, F.~McTavish, M.~Chih, A.~Atwood, R.~Johnson, M.~B. ..., and
  D.~Shah, ``A smartphone application to support recovery from alcoholism: a
  randomized clinical trial,'' {\em JAMA Psychiatry}, vol.~71, no.~5, 2014.

\bibitem{Sutton_2012_Book_ReinforcementLearning}
R.~S. Sutton and A.~G. Barto, {\em Reinforcement Learning: An Introduction}.
\newblock Cambridge, MA, USA: MIT Press, 2nd~ed., 2012.

\bibitem{Geist_2013_TNNLS_RL_valueFunctionApproximation}
M.~Geist and O.~Pietquin, ``Algorithmic survey of parametric value function
  approximation,'' {\em IEEE Transactions on Neural Networks and Learning
  Systems}, vol.~24, no.~6, pp.~845--867, 2013.

\bibitem{Grondman_2012_IEEEts_surveyOfActorCriticRL}
I.~Grondman, L.~Busoniu, G.~A.~D. Lopes, and R.~Babuska, ``A survey of
  actor-critic reinforcement learning: Standard and natural policy gradients,''
  {\em {IEEE} Trans. Systems, Man, and Cybernetics}, vol.~42, no.~6,
  pp.~1291--1307, 2012.

\bibitem{Michail_2003_JMLR_LSPI_LSTDQ}
M.~G. Lagoudakis and R.~Parr, ``Least-squares policy iteration,'' {\em J. of
  Machine Learning Research (JLMR)}, vol.~4, pp.~1107--1149, 2003.

\bibitem{fyzhu_2014_AAAI_ARSS}
F.~Zhu, B.~Fan, X.~Zhu, Y.~Wang, S.~Xiang, and C.~Pan, ``10,000+ times
  accelerated robust subset selection {(ARSS)},'' in {\em Proc. Assoc. Adv.
  Artif. Intell. (AAAI)}, pp.~3217--3224, 2015.

\bibitem{yingWang_2015_TIP_RobustUnmixing}
Y.~Wang, C.~Pan, S.~Xiang, and F.~Zhu, ``Robust hyperspectral unmixing with
  correntropy-based metric,'' {\em IEEE Transactions on Image Processing},
  vol.~24, no.~11, pp.~4027--4040, 2015.

\bibitem{fyzhu_2014_IJPRS_SSNMF}
F.~Zhu, Y.~Wang, S.~Xiang, B.~Fan, and C.~Pan, ``Structured sparse method for
  hyperspectral unmixing,'' {\em ISPRS Journal of Photogrammetry and Remote
  Sensing}, vol.~88, pp.~101--118, 2014.

\bibitem{haichangLi_2016_IJRS_LablePropagationHyperClassification}
H.~Li, Y.~Wang, S.~Xiang, J.~Duan, F.~Zhu, and C.~Pan, ``A label propagation
  method using spatial-spectral consistency for hyperspectral image
  classification,'' {\em International Journal of Remote Sensing}, vol.~37,
  no.~1, pp.~191--211, 2016.

\end{thebibliography}
}{\footnotesize \par}
\end{document}